# Political Sentiment Analysis of Persian Tweets Using CNN-LSTM Model

Mohammad Dehghani[a*], Zahra Yazdanparast[b]

[a] Electrical and Computer Engineering Department, University of Tehran, Tehran, Iran, dehghani.mohammad@ut.ac.ir
[b] Electrical and Computer Engineering Department, Tarbiat Modares University, Tehran, Iran, zahra.yazdanparast@modares.ac.ir

*A B S T R A C T*

**Sentiment analysis is the process of identifying and categorizing people's emotions or opinions regarding various topics. The analysis of Twitter sentiment has become an increasingly popular topic in recent years. In this paper, we present several machine learning and a deep learning model to analysis sentiment of Persian political tweets. Our analysis was conducted using Bag of Words and ParsBERT for word representation. We applied Gaussian Naive Bayes, Gradient Boosting, Logistic Regression, Decision Trees, Random Forests, as well as a combination of CNN and LSTM to classify the polarities of tweets. The results of this study indicate that deep learning with ParsBERT embedding performs better than machine learning. The CNN-LSTM model had the highest classification accuracy with 89 percent on the first dataset and 71 percent on the second dataset. Due to the complexity of Persian, it was a difficult task to achieve this level of efficiency. The main objective of our research was to reduce the training time while maintaining the model's performance. As a result, several adjustments were made to the model architecture and parameters. In addition to achieving the objective, the performance was slightly improved as well.**

*Keywords*— *Sentiment analysis, Persian, Machine learning, Deep learning, Twitter.*

## 1. Introduction

In text analytics, Natural Language Processing (NLP) and Machine Learning (ML) techniques are used to gain insight from unstructured text data [1]. Sentiment analysis (SA) is the study of people's opinions, attitudes, and emotions regarding certain events or topics using computational methods [2]. SA is an NLP technique used to extract and analyze emotions in text [3, 4], speech [5], and images [6]. It involves detecting emotions, classifying them (binary or multiclass), and mining opinion polarity and subjectivity [7]. Therefore, an important SA task is emotion detection (ED). In SA, the focus is mainly on specifying positive, negative, or neutral opinions, whereas in ED, the focus is on detecting a variety of emotional states from text. The implementation of ED as a SA task can be achieved using either a machine learning approach or a lexicon-based approach [2].

Over the years, people have posted a lot of content on social media like Twitter, Instagram, and Facebook and expressed their opinions, sentiments, thoughts, and attitudes towards various topics. SA can be applied to this enormous amount of textual corpora, which include real-time feedback and recommendations about events and products [8, 9]. Companies use this textual content to maintain full awareness of customer's opinion regarding their products, which is helpful in making strategic decisions and identifying limitations [10].

Twitter (which rebranded to X on July 31, 2023) is one of the most popular social media platforms and people post around 500 million tweets every day. In terms of data access, Twitter has a policy of openness and an API that can be used to collect data about any subject [11, 12]. Although tweets can be a valuable source of sentiment analysis, they present a number of challenges. Tweets are short (280 characters for free and 25,000 for paid users), informal, unstructured, and have ambiguous language with polysemy and figurative expressions [13]. The content may contain sarcasm, which conveys the intended meaning in the opposite manner. Literature is available on a variety of topics and authors. Additionally, tweets may contain Emojis or hashtags that need to be preprocessed [14].

Using tweets in Persian language poses other challenges that should be addressed. It is common in Persian to use abbreviations such as "برجام" (برنامه جامع اقدام مشترک) which means "Joint Comprehensive Plan of Action"). There are some characters in Persian alphabet that have the same pronunciation, but are written differently. These words convey negativity or positivity, which play an extremely important role in sentiment analysis. Researchers used twitter content for sentiment analysis in different domain such as healthcare [15], tourism [16], business [17], finance [18, 19], and education [20].

Political SA is no different from other domains in terms of method and implementation [21]. The importance of sentiment analysis in politics has increased in recent years for several reasons. The political diplomat must be aware of the opinions and needs of the people, as well as major issues in society [22]. Conducting surveys and polls on political topics is an extremely time-consuming and expensive endeavor. Therefore, political SA is a common method of assessing the public's attitudes regarding a political campaign [23]. SA could be used in political debates to discover the opinions of people regarding a particular election candidate or political party [24]. It is also possible to predict the election results based on political posts [22].

It has become increasingly important to analyze emotions in order to predict democratic elections [25]. Microblogging and social networking sites are considered good sources of information during the SA process since people are able to discuss their opinions freely about political issues [2].

Political parties can also benefit from sentiment analysis by assessing public opinions regarding politicians' words and actions. They can monitor the public's reactions about their promises, and determining polarities to be aware of their acceptance. This will help to fill the gap between the public and politicians and will have a significant impact on the candidate's political future.

In this paper, we aim to analyze and understand the sentiment of Persian political tweets. We considered two states. Since the first dataset has three classes, it is considered a SA job. The second dataset, on the other hand, has seven classes which are based on human emotions. For this dataset, our approach was to look at ED as a branch of SA with seven classes rather than performing SA with positive, negative, or neutral classes. Consequently, our results are more comprehensive and more comparable to human states, making them more suitable for analysis. In [23], the author performed sentiment and emotion analysis on the Gujarat Legislative Assembly Election in India. However, our approach is completely different.

In order to vectorize textual content, Bag of Words (BOW) and ParsBERT [26] were used. For prediction, we employed several machine learning classifiers as well as a deep learning method. Our results are evaluated using two Persian political datasets collected from Twitter.

In Section 2 that follows, we discuss the related work. Section 3 provides a detailed description of the proposed method. Section 4 describes the results we achieved using different models. The paper concludes with future directions in Section 5.

## 2. Related Work

The use of sentiment analysis is becoming increasingly popular in many domains, including e-learning [27], online stores [28], hotel reviews [29], and movie reviews [30]. Many researchers have used sentiment analysis to investigate events or to solve problems using Twitter data.

A sentiment analysis of three public datasets (Amazon product reviews, IMDB movie reviews, and Yelp restaurant reviews) was conducted by Mutinda et al. [31] using sentiment lexicons, N-grams, and BERT with a CNN. COVID-19 tweets were analyzed by Aslan et al. [32]. They used FastText Skip-gram to extract information, a convolutional neural network (CNN) model for feature extraction and an optimization algorithm (AOA) for feature selection. Several machine learning algorithms were used to classify tweets as positive, negative, or neutral. VADER and NRCLex were used by Ainapure et al. [33] to investigate Indian citizens' views on the COVID-19 pandemic and vaccination. They used Bi-LSTM and GRU methods to train the classification model. In the paper by AlBadani et al. [34], a methodology based on fine-tuning universal language models and SVMs was proposed to determine people's attitudes towards a particular product based on their tweets. Naseem et al. [9] present DICET, which is a transformer-based method for removing noise from tweets to enhance their quality. A BiLSTM network was applied to three datasets, including US airlines, Airlines, and Emirates airlines, to determine the sentiment of tweets. In an educational context, Misuraca et al. [35] used sentiment analysis to determine the polarity (positive/neutral/negative) of a student's feedback.

In a study by Ruz et al. [36], Bayesian networks were used to analyze sentiments during critical events. Twitter posts regarding the 2010 Chilean earthquake were collected as natural events and tweets regarding the 2017 Catalan independence referendum were collected as social movements. Neogi et al. [37] proposed a method for analyzing Indian farmers' protest tweets. They collected 18,000 tweets about the protest and applied Bag of Words and TF-IDF for word embedding. The tweets were classified using Decision Trees, Random Forests, Naive Bayes, and SVM. A study conducted by Charalampakis et al. [38] assessed how sarcastic tweets may influence electoral outcomes. Using a combination of Support Vector Machines and Neuro-Fuzzy analysis, Katta et al. [39] assessed sentiment analysis of twitter information related to political issues.

For the purpose of sentiment analysis in Persian, Shams et al. [40] present LDASA, an LDA-based sentiment analysis method, which is evaluated using data from websites about hotels, cell phones, and digital cameras. A deep learning method was proposed by Dehghani et al. [41] for detecting abusive words in Persian tweets. In order to detect the polarity of 1000 idiomatic expressions, Dashtipour et al. [42] used Decision Tree, Naive Bayes, KNN, SVM, and CNN. Using 803278 Persian tweets, Nezhad et al. [43] proposed a deep learning method based on CNN and LSTM to classify sentiments about COVID-19 vaccines. In Dastgheib et al. [44] article, structural correspondence learning (SCL) and convolutional neural networks (CNN) are combined in order to analyze the polarity of 26K Persian sentences obtained from user opinions about digital products. In subject of Persian political tweets, our previous work [45] was the first study. We used a CNN-BiLSTM model which scored the accuracy of 0.889 on first dataset and 0.710 on second dataset.

## 3. Methodology

In this section, we describe our proposed method for analyzing Persian political tweets. Our approach is illustrated in Figure 1. We preprocessed the input data and then used BOW and ParsBERT to convert textual information into numerical vectors. As Bow is a traditional method, ParsBERT is a fine-tuned version of BERT that takes the semantics into account. Several algorithms were used to classify tweets polarity, namely Gaussian Naive Bayes, Gradient Boosting, Logistic Regression, Decision Tree, Random Forest, as well as CNN-LSTM. The BOW was mostly used with traditional machine learning algorithms and the ParsBERT output was used as an input for the deep learning model. Lastly, we evaluate every algorithm based on accuracy, precision, recall and F1-score.

This paper covers a wide range of traditional machine learning algorithms and a deep learning model that provides a comparison between the performance of different models.

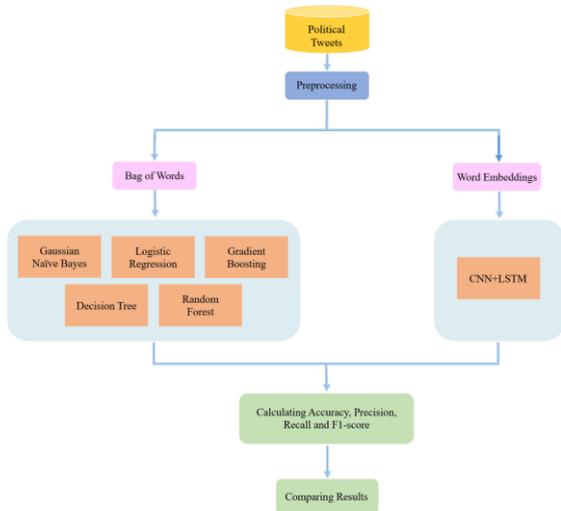

Figure. 1. Our proposed method.

### 3.1. Dataset

In this study, we analyzed two Persian political tweets datasets to analyze the polarity of people's opinions about Iranian politicians. The first dataset divided into three categories based on positive, negative, and neutral polarity. In the second dataset, there are seven different emotion categories, including happy, sad, hopeless, hopeful, angry, apathy, and others. A sample of each class presented in the second dataset is shown in Table 1.

### 3.2. Data Preprocessing

Data preprocessing is the process of preparing raw data before it is used to build Machine Learning models which involve several steps. As tweets are sometimes posted in a variety of ways, there is a lot of inconsistency and redundancy, which makes data cleaning necessary. The following preprocessing steps were applied:

- Removing all HTMLs and URLs: It should be noted the data was collected from Twitter, a social media platform where users can post images, videos and links. Typically, HTML tags are used to present this type of content, which is collected during the crawling process. We should remove these HTML tags and other URLs in order to clean the data [46].
- Removing repetitive characters in a word, e.g. "خوووووب" is replaced with "خوب" (Persian word for good).
- Additional vocabulary pruning was carried on to avoid non-informative terms, the so-called stopwords [47].
- Emojies were replaced with their textual equivalent using Regex.
- Stemming is applied to extract the base form of the words.
- Detecting incorrect spellings and correcting them.

Table 1: Samples of second dataset.

| Tag | English | Persian |
|---|---|---|
| Happy | In a country where logic is rare on the part of the statesmen, logical remarks by president of the central bank is hilarious! #Currency #dollar | تو مملکتی که شنیدن حرف حساب از مسئولان کیمیاست، شنیدن چارکلام حرف حساب از رییس بانک مرکزی چقدر می چسبه! #ارز #دلار |
| Sad | The factory where my father works is being closed due to the #strike of truck driver. | کارخونه‌ای که بابام توش کار میکنه داره بسته میشه سر #اعتصاب-کامیونداران |
| Hopeful | Iranians are engaged in various professions abroad. Most of them, from scientist to worker, are elite in their profession. They have been vagabonds for so long and nowhere feels like their homeland. The majority will return and we will rebuild the country as a united nation. | ایرانیان در خارج به حرفه‌های مختلف مشغولند. حداکثرشان، از دانشمند تا کارگر در حرفه خود نخبه‌اند. 40 سال است در کشورهای خارجی خانه بدوشند. هر کجا هستند میهنشان نیست. اکثریت برمی‌گردند و دوباره ایران را با اتحاد ملت ایران در داخل می‌سازیم. |
| Hopeless | Why should I stay and not emigrate?! | چرا باید بمونم و مهاجرت نکنم؟! |
| Angry | They gloriously and publicly announce the way they spend the money of labor children and orphans. They consider us so fool and inferior that they so mock us. Till when should we pretend to be asleep | با افتخار میگن نون بچه‌های کار و کودکان یتیم ایران را کجا و چگونه هزینه کردند. انقدر ما رو احمق و ذلیل میدونن که این چنین به ریشمان میخندند تا کی خودمون رو به خواب بزنیم |
| Apathy | Political news is none of my business. | اخبار سیاسی ب من چ اخه |
| Others | Tomorrow is election day | فردا روز انتخاباته |

### 3.3. Feature Representation

We used BOW and ParsBERT techniques to convert textual tweets into numeric format. BOW is a simple, yet powerful, method of vectorizing text. This method counts the number of times each word appears in a document and does not consider its position. BOW generates a document-term matrix (DTM) that describes the frequency of terms within a collection of documents. DTM is used as input for machine learning classifier.

ParsBERT is a monolingual language model based on Google's BERT architecture with the same configurations as BERT-Base. A large Persian corpus (with more than 2M documents) has been used to pre-train this model. There are a variety of writing styles on the corpus, as well as many subjects (e.g., scientific, novels, news). With ParsBERT, the semantics and meaning of the Persian sentences are taken into account [48]. ParsBERT is intended to recognize the nuances and complexities of the Persian language to be more efficient with NLP tasks.

### 3.4. Model Building

*Gaussian Naïve Bayes*

A naive Bayes classification algorithm is based on Bayes' Theorem, where features are assumed to be independent of each other. Bayesian networks assume features are independent. The algorithm calculates the probability of an item occurring, and then assigns the document to the class which has the highest posterior probability [49]. By using the Bayes theorem, we can calculate the posterior probability P(A|B). The equation is as follows:

$$P(A|B) = \frac{P(B|A)\,P(A)}{P(B)} \quad (1)$$

where P(A|B) is the posterior probability of the target class, P(A) is the class prior probability, P(B|A) is the likelihood which is the probability of predictor of given class and P(B) is the predictor prior probability.

The Gaussian Naive Bayes classification algorithm is an extension of the Naive Bayes classification algorithm. Gaussian Naive Bayes assumes that data follows a normal distribution and each parameter (also known as feature or predictor) has an independent ability to predict the output variable. Combining all predictions yields the final prediction, which returns a probability for the dependent variable to be classified in each group. As a result, the final classification is assigned to the group with the highest probability. In Equ (2), a Gaussian Naive Bayes calculation is shown. The maximum likelihood is used for estimating $\sigma_B$ and $\mu_B$ [50].

$$p(A_i|B) = \frac{1}{\sqrt{2\pi\sigma_B^2}} e^{\left(-\frac{(A-\mu_B)^2}{2\sigma_B^2}\right)} \quad (2)$$

*Gradient Boosting*

The gradient boosting method is one of the types of ensemble learning. Ensemble learning combines a group of weak learners to create a strong learner [51]. In ensemble boosting, models are produced sequentially by iteratively minimizing the error of earlier learned models. In gradient boosting, an ensemble of weak learners is used to improve the performance of a machine learning model. In classification, the final result is calculated as the class that receives the majority of votes from weak learners. In gradient boosting, weak learners work sequentially. The goal of each model is to improve on the error of the previous model. It is different from bagging, in which multiple models are fitted on subsets of data in parallel [52]. An example of bagging is Random Forest.

*Logistic Regression*

A logistic regression model, despite its name, is a classification model, not a regression model. It is a classification model that is simple to implement and performs well with linearly separable classes [53]. The logistic regression algorithm is based on logistic function. Logistic functions are sigmoid curves with the following equation:

$$f(x) = \frac{M}{1 + e^{-k(x-x_0)}} \quad (3)$$

where e represents Euler's number, $x_0$ represents the midpoint of the sigmoid curve, M represents its maximum value, and s represents its steepness.

Logistic Regression converts categorical dependent variables into probability scores. These scores are then used to measure the relationship between the categorical dependent variable and the continuous independent variable. A logistic regression can be classified into binary and multinomial types. A multinomial logistic regression is used when the labels contain multiple values [54, 55].

*Decision Tree*

In machine learning, decision trees are used for classification and follow an "if/then" logic. During the development of a decision tree, the tweet dataset is the root note and the decision nodes represent the features of the dataset. Decision nodes are split into branches and create child nodes. The branches are actually the decision rules. The output of a decision tree is represented by leaf nodes, which have no branches [37]. For the first dataset, leaf nodes may be positive, negative, or neutral, and for the second dataset, they may be happy, sad, hopeless, hopeful, angry, apathy, or others. A decision tree provides an explainable model and is simple.

*Random Forest*

A random forest is a collection of decision trees that can be used for both regression and classification. Each decision tree classifier receives a part of the data, produces an output, and the final result is decided by voting. Random Forest is an ensemble algorithm that uses bagging approach in order to determine voting results. In comparison with decision trees, the random forest classifier has the advantage of eliminating the possibility of overfitting [55, 56].

*CNN-LSTM*

Convolutional neural networks (CNNs) are a type of artificial neural networks that learn directly from data. CNN is a feedforward network that can extract features from data with convolutional structures [57]. As a result of the convolution layer, the input data is filtered and a feature map is created, which illustrates the particular attributes associated with the data points [58].

Long short-term memory (LSTM) is an improved model of recurrent neural networks (RNNs) that solves the gradient vanishing or exploding problem [59]. LSTM models have three gates: the input gate, the forget gate, and the output gate. Input gate determines which information is added to memory cell. Forget gate determines which information should be removed from memory cell and which information should be retrained. It enables LSTM to solve the gradient vanishing or exploding problem. Output gate is responsible for calculating the output. LSTM is capable of remembering information for a long time, which makes them suitable for text classification [60]. Equ (4)-(7) explain the LSTM gates and cells [61].

Input gate $\quad In_t = \sigma\,(W_{in}[hs_t - 1].x_t + b_{in}) \quad (4)$

| Forget gate | $f_t = \sigma(W_f[hs_t - 1].x_t + b_f)$ | (5) |
| Output gate | $f_o = \sigma(W_o[hs_t - 1].x_t + b_o)$ | (6) |
| Memory cell | $C_t = tanh(W_c[hs_t - 1].x_t + b_e)$ | (7) |

CNN is capable of detecting local features and deep features from text, and LSTM is capable of processing sequential input. Figure 2 shows our proposed deep learning architecture that combines CNN and LSTM. First, we used ParsBERT for word embedding, resulting in text vectorization, which can be used as input to a neural network. For feature extraction, One-dimensional Convolution (Conv1D) is used. The next layer is the max-pooling which reduces the network parameters, resulting in a faster training process and the ease of handling overfitting problems [62]. To train the classification model, a LSTM layer was added. A spatial dropout layer follows with a dropout rate of 0.1, which sets 10 percent of input units to zero during training. As a final layer, the dense layer is used to generate predictions.

Although the proposed architecture in the previous work [45] achieved good results, the train and inference times were high and the method was not fast as compared to the machine learning methods. As part of this work, we sought to make changes that would increase speed while maintaining accuracy. TF-IDF was replaced by BOW for vectorizing input for machine learning algorithms. BOW is a simpler and faster method. However, the major changes have been made to the deep learning model, which had produced better results. There are two areas where we could make changes: the architecture of the model and the parameters of the model.

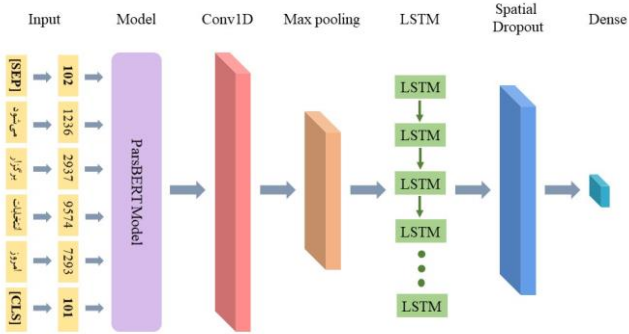

Figure. 2. Architecture of the deep learning model.

**Model architecture:** The model's architecture was changed by reducing the size of the layers. The convolution layer size has been reduced from 64 to 32, and the maxpooling size has been reduced from 8 to 2. Additionally, LSTM was used rather than BiLSTM. LSTM-based models are slightly preferred over BiLSTM-based models since they require a greater amount of computational time to train [63]. Furthermore, dropout layer was replaced by spatial drop out. With spatial drop out, an entire 1D vector is dropped out at the specified rate, preventing the model from overfitting and ensuring better generalization [64].

**Model parameters:** There are five different parameters used in the neural network, each of which can have a different value. We considered 5, 10, 20, 30, 50, and 100 for epochs and 2, 4, 8, 16, 32, and 64 for batch size. In this experiment, the learning rate was set at 0.1, 0.01, 0.001, 0.0001, 0.00001, and 0.000001. We selected accuracy and categorical cross-entropy for the loss and Adam and SGD as the optimizer. These values are summarized in Table 2. Therefore, the most optimal parameters were determined by greedy search. The greedy algorithm attempts to arrive at a globally optimal solution by making a locally optimal choice [65].

Table 2: Model parameters for greedy search.

| Parameter | Values |
| --- | --- |
| Epochs | 5, 10, 20, 30, 50, 100 |
| Batch size | 2, 4, 8, 16, 32, 64 |
| Learning rate | 0.1, 0.01, 0.001, 0.0001, 0.00001, 0.000001 |
| Loss | Accuracy, Categorical Cross-Entropy |
| Optimizer | Adam, SGD |

## 4. Results and Discution

### 4.1. Evaluation Metrics

To evaluate the performance of each model, we used accuracy, precision, recall and F1-score measures. They are calculated based on confusion matrix which has four values:
- True Positives (TP): The number of positive tweets that have been classified correctly.
- True Negatives (TN): The number of negative tweets that were correctly classified.
- False Positives (FP): The number of tweets that have been incorrectly classified as positive.
- False Negatives (FN): The number of tweets that have been incorrectly classified as negative.

**Accuracy** is defined as the proportion of true results among all the cases examined. It can be used for binary as well as multiclass classification problems.

$$Accuracy = \frac{TP + TN}{TP + FP + TN + FN} \quad (8)$$

**Precision** indicates what proportion of predicted positives are actually Positive.

$$Precission = \frac{TP}{TP + FP} \quad (9)$$

**Recall** measures the proportion of Positives that are correctly classified.

$$Recall = \frac{TP}{TP + FN} \quad (10)$$

**F1-score** is a number between zero and one that represents the harmonic mean of precision and recall.

$$F1\_score = \frac{2 * Precission * Recall}{Precission + Recall} \quad (11)$$

### 4.2. Evaluation

This section presents results of each dataset. Parameters used in training of deep learning model are shown in Table 3, which are the result of greedy search. Loss function is calculated based on Categorical cross-entropy. Adam algorithm is used to optimize the training model. Learning rate is set to 0.0001 and the neural networks were trained for 10 epochs with a batch size equal to 64. Datasets were split 70:30, with 70% of the data used for training the models and 30% for evaluating their performance.

Table 4 shows the results of first dataset. Gradient Boosting shows the best results in terms of precision, equal to 90%. The performance of Random Forest was more accurate when compared to other machine learning methods, with accuracy, recall, and F1-score equal to 86%. CNN-LSTM has the highest performance with accuracy and recall of 89% and F1-score of 88%. Therefore, the overall result of deep learning is better than the result of machine learning.

The results of the sentiment analysis of the second dataset are presented in Table 5. Among all classifiers, CNN-LSTM has the best accuracy, precision, recall, and F1-score, which is 71%. The performance of logistic regression is better than other machine learning models with accuracy, precision, recall, and F1-score of 66%.

According to the results, deep learning performed better than machine learning in both datasets. Input for the machine learning algorithm is derived from BOW, which constructs a document vector based on words without considering semantics. ParsBert is a fine tuning of BERT embeddings that take context into account, and this result is used as input to CNN-LSTM models. Furthermore, the CNN layer extract features that are useful to train LSTM.

As we can see from Tables 4 and 5, the models perform better when analyzing sentiments for three classes compared to seven classes.

Table 3: Model parameters

| Parameter | Value |
|---|---|
| Epochs | 10 |
| Batch size | 64 |
| Learning rate | 0.0001 |
| Loss | Categorical cross-entropy |
| Optimizer | Adam |

Table 4: First dataset results.

| Model | Accuracy | Precision | Recall | F1-score |
|---|---|---|---|---|
| Gaussian Naïve Bayes | 0.619 | 0.669 | 0.619 | 0.591 |
| Decision Tree | 0.761 | 0.771 | 0.761 | 0.766 |
| Gradient Boosting | 0.765 | **0.901** | 0.765 | 0.805 |
| Random Forest | 0.869 | 0.866 | 0.868 | 0.865 |
| Logistic Regression | 0.846 | 0.875 | 0.846 | 0.855 |
| CNN-LSTM | **0.892** | 0.888 | **0.892** | **0.888** |

Table 5: Second dataset results.

| Model | Accuracy | Precision | Recall | F1-score |
|---|---|---|---|---|
| Gaussian Naïve Bayes | 0.632 | 0.636 | 0.632 | 0.632 |
| Decision Tree | 0.594 | 0.595 | 0.594 | 0.594 |
| Gradient Boosting | 0.64 | 0.655 | 0.64 | 0.642 |
| Random Forest | 0.653 | 0.669 | 0.653 | 0.656 |
| Logistic Regression | 0.668 | 0.669 | 0.668 | 0.668 |
| CNN-LSTM | **0.714** | **0.717** | **0.714** | **0.713** |

Our main objective was to decrease training and inference times as compared to previous models [45] without compromising model accuracy. A comparison of the results is presented in Table 6. Using the new lighter model, we have improved SA results in both datasets. The accuracy of the first dataset, which has three classes, improved by 0.3%. Accuracy improved by 0.4% in the second dataset, which has seven classes. The improvement is not significant, but our primary goal was to reduce computation time while maintaining the model's performance. According to the results, the goal was attained.

The hybrid CNN-LSTM model that we have proposed in this paper has proven to be highly effective by providing the best results. The results obtained in this paper indicate that combining deep learning algorithms effectively is a promising approach to creating efficient political SA and ED systems.

Table 6: Comparing results.

| | Model | Accuracy | Precision | Recall | F1-score |
|---|---|---|---|---|---|
| First Dataset | [45] | 0.889 | - | - | 0.887 |
| | CNN+LSTM | 0.892 | 0.888 | 0.892 | 0.888 |
| Second Dataset | [45] | 0.710 | - | - | 0.700 |
| | CNN+LSTM | 0.714 | 0.717 | 0.714 | 0.713 |

## 5. Conclusion

We present several machine learning methods (Gaussian Naive Bayes, Gradient Boosting, Logistic Regression, Decision Trees and Random Forests) and deep learning (CNN-LSTM) for sentiment analysis of Persian political tweets. We vectorized tweets with BOW for machine learning, and ParsBERT for deep learning. Models were evaluated and compared using two datasets from twitter. There are three classes in the first dataset and seven classes in the second dataset. In the first dataset, CNN-LSTM performed better with the highest accuracy (89%), recall (89%), and F1-score (88%). As well, it has the best performance in all evaluation metrics for the second dataset with accuracy, precision, recall, and F1-score of 71%. Additionally, we aimed to reduce the training time and increase the speed of the model, which was accomplished.

We developed a hybrid model that combines CNN with LSTM in order to take advantage of the benefits of both layers. The model was also evaluated against several known machine learning algorithms. Based on several evaluation metrics, our CNN-LSTM model is found to provide the best performance.

In the future, other features can be considered, such as hashtag frequency. In order to gain better results, using multimodal approaches and combining textual tweets with other types of data such as speech and video can be beneficial.

Considering that fine-tuning the Bert model is state of the art in different tasks, it would be interesting to see how it performs on other dataset. Our proposed method can be applied to other political datasets in other languages, such as English, and the results can be compared.

## Declarations

### *Funding*
This research did not receive any grant from funding agencies in the public, commercial, or non-profit sectors.

### *Authors' contributions*
First Author: Study design, acquisition of data, implementation of the computer code and supporting algorithms, interpretation of the results;

Second Author: Writing and editing the manuscript;
All authors read and approved the final manuscript.

### *Conflict of interest*
The authors have no conflicts of interest to declare.